%% file: latex/acl_latex.tex
\pdfoutput=1

\documentclass[11pt]{article}

\usepackage[final]{acl}
\usepackage{booktabs}
\usepackage{tabularx}
\usepackage{listings}
\usepackage{amsmath,amssymb}   
\usepackage{times}
\usepackage{latexsym}

\usepackage[T1]{fontenc}

\usepackage[utf8]{inputenc}

\usepackage{microtype}

\usepackage{inconsolata}

\usepackage{graphicx}
\usepackage{subfigure}
\usepackage{booktabs}

\usepackage{amsmath}
\usepackage{amsthm}
\usepackage{enumitem}
\usepackage{color,amsfonts}
\usepackage{makecell}
\usepackage{multirow,array}
\usepackage{tikz}
\usepackage{pgfplots}
\usepackage{amssymb,mathtools}
\usepackage[linesnumbered,ruled,vlined]{algorithm2e}
\usepackage{caption}
\usepackage{cleveref}
\usepackage[table]{xcolor}
\usepackage{arydshln}
\usepackage{amssymb}

\title{Evidence-based Distributional Alignment for Large Language Models}

\author{Viet-Thanh Pham$^1$, Lizhen Qu$^1$, Zhuang Li$^2$, Gholamreza Haffari$^1$ \\ $^1$Department of Data Science \& AI, Monash University, Australia
\\ $^2$Royal Melbourne Institute of Technology, Australia
\\ {\small \texttt{thanh.pham1@monash.edu, lizhen.qu@monash.edu, zhuang.li@rmit.edu.au,}}
\\ {\small \texttt{ gholamreza.haffari@monash.edu}}}

\begin{document}
\maketitle
\begin{abstract}

\input{latex/Abstract}
\end{abstract}
\input{latex/Introduction}

\input{latex/ProposedMethod}
\input{latex/Experiments}
\input{latex/RelatedWorks}
\input{latex/Conclusions}
\input{latex/Limitations}

\bibliography{custom}
\clearpage
\appendix

\input{latex/Appendix}

\end{document}

%% file: latex/Abstract.tex
Distributional alignment enables large language models (LLMs) to predict how a target population distributes its responses across answer options, rather than collapsing disagreement into a single consensus answer. However, existing LLM-based distribution prediction is often unstable and degrades under cultural and domain shift. Token score-based estimates can change with minor option wording or formatting, response sampling-based estimates are expensive and sensitive to prompts and decoding settings, and directly generated distributions are frequently miscalibrated.

We propose Evi-DA, an evidence-based alignment technique that improves the fidelity and robustness of LLM-based distribution estimation under domain and cultural shift. Given a target country and a multiple-choice question, Evi-DA retrieves related World Values Survey items and their answer distributions, predicts a coarse Welzel value signature for each option, and infers the country-conditioned answer distribution in a structured format. We train the LLMs using a two-stage pipeline, where reinforcement learning optimizes survey-derived rewards that encourage accurate intermediate value predictions, faithful final distributions, well-formed structured outputs, and reduced cultural bias. Across in-domain and out-of-domain benchmarks and multiple open-source backbones, Evi-DA reduces Jensen-Shannon divergence between predicted and gold distributions relative to strong baselines, with average relative improvements of up to 44\%.


%% file: latex/Introduction.tex
\section{Introduction}

Motivated by pluralistic alignment \cite{pluralistic, feng-etal-2024-modular}, distributional alignment for large language models (LLMs) aims to make model outputs reflect population-level diversity in values and opinions rather than collapsing it into a single consensus. It supports LLM-based applications such as comparing opinion gaps across groups, stress testing downstream decisions under demographic heterogeneity, and building population-level social simulations \cite{bougie2025citysimmodelingurbanbehaviors, Argyle_2023}. These applications are important in real-world settings, where there is rarely a single correct answer and meaningful disagreement arises across communities. 
A central task for LLMs under distributional alignment is \emph{distribution prediction}, where a system estimates a distribution over answer options that approximates how a target population or demographic group would respond \cite{meister-etal-2025-benchmarking}. This is useful for computational social science, such as estimating how a group might answer multiple choice questions before a full-scale survey \cite{pham-etal-2025-surveypilot, opinionqa}. 


Existing LLM-based distribution prediction methods are often fragile. Token-score approaches \cite{meister-etal-2025-benchmarking} can be unreliable because early token log-probability scores do not consistently correspond to selecting a particular option, especially when options are multiword or formatted differently. Repeated response sampling \cite{ferraro2024agentbasedmodellingmeetsgenerative, chuang-etal-2024-simulating} can approximate human distributions, but it is computationally expensive and highly sensitive to prompts and decoding settings, making it hard to know whether the differences we see come from real disagreement among people or are just caused by the way the model is asked or sampled. Directly prompting LLMs to output a verbalized distribution \cite{wang2024calibratingverbalizedprobabilitieslarge, meister-etal-2025-benchmarking} is convenient but often miscalibrated and prone to surface-form anchoring, producing plausible-looking distributions that drift from real survey behavior. These weaknesses are amplified under domain and cultural shift, where skews in pretraining and post-training data can systematically misalign predictions for underrepresented groups and reduce robustness outside training-like regimes \cite{durmus2024towards, naous-etal-2024-beer, chiu-etal-2025-culturalbench}.


To address these limitations, we propose Evi-DA, an evidence-based distributional alignment framework that improves LLM prediction fidelity and robustness under domain and cultural shift by reasoning through an interpretable intermediate representation grounded in human values. The key idea is to route prediction through Welzel’s human values framework, a compact latent space that captures cross-cultural variation via a small set of value dimensions \cite{WelzelWVS_SecularEmancipative_Syntax}. Evi-DA retrieves relevant evidence from large-scale survey data, using the World Values Survey (WVS) as a high-coverage source of empirical distributions across countries and demographic attributes \cite{Inglehart2022WVS}. Given a new question, it predicts how each answer option maps to coarse value levels, then combines these predicted value signatures with the retrieved evidence to infer the demographic-conditioned answer distribution. This decomposition makes intermediate reasoning steps explicit, reduces sensitivity to option surface forms, and yields a more inspectable and consistent prediction process. We train the LLMs with a two-stage reinforcement learning procedure based on Group Relative Policy Optimization (GRPO; \cite{shao2024deepseekmath}), using survey-derived rewards that encourage accurate intermediate value predictions, faithful final distributions, structured outputs, and reduced cultural bias.




We evaluate Evi-DA on \emph{country-conditioned distribution prediction} for multiple-choice survey questions under in-domain and out-of-domain shifts. Experiments cover six open-source LLM backbones and seven strong baselines, such as token log-probability estimation, large-scale opinion sampling, verbalized distribution generation, and agentic/multi-agent approaches. Evi-DA consistently lowers Jensen--Shannon divergence (JSD) between predicted and gold answer distributions, reaching 0.16 in-domain and 0.23 out-of-domain on average. This yields up to 44\% relative JSD reduction over token log-probabilities and 21\% over the best baseline under shift. GRPO-based reinforcement learning typically provides additional gains by explicitly rewarding intermediate value-signature accuracy and faithful final distributions.

In summary, this work makes two contributions:
\begin{itemize}
    \item We introduce Evi-DA, an evidence-based distributional alignment framework that retrieves evidence from WVS and uses Welzel value signatures as a transferable intermediate representation, enabling interpretable two-stage inference for distribution prediction.
    \item We provide a comprehensive evaluation on 2{,}403 country-conditioned questions (2{,}203 in-domain; 200 out-of-domain) across six LLM backbones and seven baselines, demonstrating consistent improvements and further gains from GRPO training objectives.
\end{itemize}

%% file: latex/ProposedMethod.tex
\section{Method}
\label{sec:method}

\input{Figures/evi_da}

\subsection{Overview}
\Cref{fig:evi_da} illustrates Evi-DA, our evidence-based framework for distributional alignment. We adopt the pluralistic alignment view~\citep{pluralistic}: rather than returning a single ``best'' answer, an LLM should estimate how a target population distributes its responses across the available options. Concretely, we study \emph{distribution prediction conditioned on country and demographic attributes} for survey-style multiple-choice questions \cite{pham-etal-2025-surveypilot, Argyle_2023}. Given a prompt $x$ with options $\mathcal{O}_x=\{o_1,\ldots,o_K\}$ and a demographic group $g$ (for example, country, gender, age interval), the goal is to estimate the group-level answer distribution $P(\cdot \mid x,g)$ over $\mathcal{O}_x$ and output it in a fixed JSON schema that assigns a probability to each option.

Evi-DA improves distribution prediction fidelity and robustness under domain and cultural shifts by decomposing country and group-conditioned distribution prediction into five modules: (i) Welzel-based value representation (\S\ref{sec:welzel}), (ii) evidence bank construction from WVS (\S\ref{sec:evidence}), (iii) evidence retrieval for a new question (\S\ref{sec:retrieval}), (iv) two-stage inference with structured intermediate representations (\S\ref{sec:twostage}), and (v) GRPO-based reinforcement learning (\S\ref{sec:grpo}).


\subsection{Welzel values as a latent representation} \label{sec:welzel}
Welzel’s framework summarizes cross-cultural variation in public opinion with two higher-level dimensions: \emph{Secular Values} and \emph{Emancipative Values}. Secular values reflect orientation toward secular-rational worldviews rather than traditional authority and sacred norms, while emancipative values capture support for autonomy, equality, freedom of choice, and participatory voice (Appendix~\ref{appendix:welzel}). Crucially for distributional alignment, this yields a compact, interpretable coordinate system that is comparable across domains.

We represent each respondent \(i\) by a Welzel value profile vector \(\mathbf{z}_i \in [0,1]^D\), where \(D=8\) indexes the ordered sub-indices:
\(\mathbf{z}_i=\big(z_i^{(1)},\ldots,z_i^{(8)}\big)\).
The first four components \(z_i^{(1:4)}\) correspond to secular-side values, and the last four \(z_i^{(5:8)}\) correspond to emancipative-side values. We compute \(\mathbf{z}_i\) by applying Welzel’s published index construction to each respondent’s answers and metadata across all WVS items (details in Appendix~\ref{appendix:welzel}).

\subsection{Evidence bank construction from the World Values Survey} \label{sec:evidence}
We build an evidence bank from WVS dataset. WVS contains the human-annotated value profile vector \(\mathbf{z}_i\) for each of its data samples. For each demographic group $g$ and WVS item $q$, it stores (i) the empirical answer distribution of $g$ for $q$, and (ii) a value-based signature for each answer option, derived from the respondents who selected it. This produces group-specific calibration examples that can be retrieved to ground distribution prediction for new questions.

Let $\mathcal{I}_g$ be the set of respondents in group $g$, and let $\mathcal{I}_{g,q}\subseteq \mathcal{I}_g$ denote those with an observed answer to item $q$. For option $v$ of item $q$, define
\begin{equation}
\mathcal{I}_{g,q,v} = \{ i \in \mathcal{I}_{g,q} : y_i^q = v \},
\end{equation}
where $y_i^q$ is respondent $i$’s choice. The empirical answer distribution is
\begin{equation}
p_g^q(v) = \frac{1}{|\mathcal{I}_{g,q}|}\sum_{i\in \mathcal{I}_{g,q}} \mathbb{I}\!\left[y_i^q=v\right].
\label{eq:group_dist}
\end{equation}
To connect options to values, we summarize the value profiles of respondents who select each option. For option $v$ of item $q$, we compute the mean Welzel vector
\begin{equation}
\boldsymbol{\mu}_{g,q,v} = \frac{1}{|\mathcal{I}_{g,q,v}|}\sum_{i\in\mathcal{I}_{g,q,v}} \mathbf{z}_i.
\label{eq:option_mean_value}
\end{equation}
Because continuous value scores are often noisy as LLM context, we discretize each sub-index into \emph{low}, \emph{medium}, or \emph{high} (LMH). Let $0<\tau_1<\tau_2<1$ and define, for $a\in[0,1]$,
\begin{equation}
h(a)=
\begin{cases}
\text{low}, & a < \tau_1,\\
\text{medium}, & \tau_1 \le a < \tau_2,\\
\text{high}, & a \ge \tau_2.
\end{cases}
\end{equation}
Applying $h$ component-wise to $\mu_{g,q,v}$ yields a categorical signature $\mathbf{c}_{g,q,v}$:
\begin{equation}
\mathbf{c}_{g,q,v} = h(\boldsymbol{\mu}_{g,q,v}) \in \{\text{low},\text{medium},\text{high}\}^{8}.
\label{eq:option_lmh}
\end{equation}
Intuitively, $\mathbf{c}_{g,q,v}$ captures the value profile that tends to select option $v$ within group $g$. The two hresholds $\tau_1,\tau_2$ are set in Appendix~\ref{appendix:welzel_mapping}.

\subsection{Retrieving evidence for input question}
\label{sec:retrieval}

Given a new question \(x\), we retrieve a small set of related questions from the evidence bank to supply grounded exemplars from the same demographic group. Each survey item \(q\) is associated with a textual description (question text and optional instruction) and a set of labeled options. We represent item \(q\) by a string \(t_q\) formed by concatenating its question text with any available instruction, and analogously represent the input question by \(t_x\). Retrieval scores each candidate item by cosine similarity in a vector space induced by a text encoder \(\phi(\cdot)\) (e.g., BERT):
\begin{equation}
s(q \mid x)=\cos\!\big(\phi(t_q), \phi(t_x)\big).
\label{eq:retrieval_score}
\end{equation}
We rank items by \(s(q \mid x)\) and take the top-\(K\) candidates (\(K=10\)), then apply a data-support constraint, retaining only questions with sufficient valid responses in group \(g\), i.e., \(|\mathcal{I}_{g,q}|\ge N_{\min}\). This support filter is critical because the option signatures in Eq.~\eqref{eq:option_lmh} become unstable when computed from too few respondents, and unstable signatures can mislead the subsequent inference. The resulting retrieved evidence set is \(\mathcal{Q}_x = \{q_1,\ldots,q_M\}\) with \(M\le K\), where each \(q \in \mathcal{Q}_x\) is both textually related to \(x\) and statistically reliable for group \(g\). For each retrieved question, we provide the pair \(\big(p_g^q, \{\mathbf{c}_{g,q,v}\}_v\big)\) as evidence. Retrieval is not intended to find identical or near-duplicate questions. Its role is to supply calibration examples that expose how LMH value signatures map to empirical distributions within the target group. Even when \(x\) comes from an unseen domain, retrieval often yields partially related items that share preference structure (e.g., questions about openness, autonomy, social tolerance, lifestyle, or identity), and these examples constrain the model's distribution in a way that pure prompting cannot.

\subsection{Two-stage inference with structured intermediate representations}
\label{sec:twostage}

The inference procedure decomposes distribution prediction into two subtasks. This decomposition is motivated by the observation that predicting an entire distribution for a new question requires both semantic interpretation of options and calibration to a population. LLMs are strong at interpreting text but unreliable at population calibration without evidence, while the evidence bank provides population calibration but cannot interpret unseen options. We therefore first map the new options into Welzel's value space, and only then predict a distribution conditioned on the retrieved evidence and the predicted option signatures.

\paragraph{Stage A: option value profiling.}
For a new question \(x\) with options \(\mathcal{O}_x\), the model predicts an LMH signature for each option:
\begin{equation}
\hat{\mathbf{c}}_{x,o} \in \{\text{low},\text{medium},\text{high}\}^{8} \qquad \forall o \in \mathcal{O}_x.
\label{eq:pred_option_lmh}
\end{equation}
The input context includes the group profile \(\mathbf{c}_g\) and the retrieved evidence items
\(\{(p_g^q,\{\mathbf{c}_{g,q,v}\}_v): q \in \mathcal{Q}_x\}\),
i.e., each retrieved question's observed distribution together with its option signatures (prompt template provided in Appendix~\ref{appendix:prompts}, \Cref{listing-lmh-prediction}). \textit{In the prompt, each retrieved \(\mathbf{c}_{g,q,v}\) is serialized as an ordered list of eight LMH string labels, and the output \(\hat{\mathbf{c}}_{x,o}\) is in the same form.} Conceptually, Stage A uses linguistic cues from option text to infer whether selecting the option is more consistent with high/low levels of each sub-index, while using the retrieved evidence as a prior over plausible patterns. The predicted signatures provide an interpretable \textit{intermediate representation} whose primary role is to make the second-stage distribution prediction more stable and controllable.


\paragraph{Stage B: distribution prediction conditioned on value evidence.}
Given the predicted option signatures from Stage A and the retrieved evidence \(\{(p_g^q,\{\mathbf{c}_{g,q,v}\}_v): q \in \mathcal{Q}_x\}\), the LLM predicts a distribution \(\hat{P}(\cdot \mid x,g)\) over \(\mathcal{O}_x\). The input includes \(\mathbf{c}_g\), the retrieved questions with their observed distributions and option signatures, the new question text, and the predicted signatures (prompt template provided in Appendix~\ref{appendix:prompts}, \Cref{listing-distribution-prediction}). Both stages are implemented as structured JSON generation tasks with explicit schemas. This is a core part of the method rather than an implementation detail: it forces the model to explicitly commit to an LMH vector per option and to assign probability mass to every option, making failure modes observable and enabling direct automatic scoring for reinforcement learning with GRPO during training.

\subsection{Reinforcement learning with GRPO}
\label{sec:grpo}

The two-stage pipeline can be applied in a zero-shot manner, but it is also well-suited to reinforcement learning because it exposes intermediate and final supervision signals that can be computed from survey data without manual annotation. We apply GRPO by treating the LLM as a policy that emits the structured outputs for Stage A and Stage B.

We construct training episodes directly from WVS as follows. For a demographic group \(g\), we \textsc{(i)} sample a WVS item \(q\) and treat it as a pseudo ``unseen'' input \(x\) (using its question text and options); \textsc{(ii)} construct an evidence set \(\mathcal{Q}_x\) from an evidence split (excluding \(q\)); and \textsc{(iii)} use WVS-derived supervision for \(q\) in group \(g\): the empirical answer distribution \(p_g^q\) from Eq.~\eqref{eq:group_dist} as the ground-truth target for Stage B, and the option signatures \(\{\mathbf{c}_{g,q,v}\}_v\) from Eq.~\eqref{eq:option_lmh} as ground truth for supervising Stage A. This yields scalable, self-contained supervision across many question--group pairs.

We define four reward components for the two tasks and schema validity. The \textbf{first reward} measures correctness of LMH prediction in Stage A. Let \(\hat{\mathbf{c}}_{x,o}\) be the model's predicted LMH vector for option \(o\). When \(x\) is a held-out survey item, we denote the corresponding survey-derived ground-truth signature by \(\mathbf{c}^{*}_{x,o}\). We define the LMH accuracy reward as average per-dimension exact match:
\begin{equation}
R_{\text{LMH}}=\frac{1}{|\mathcal{O}_x|D}\sum_{o\in\mathcal{O}_x}\sum_{d=1}^{D}
\mathbb{I}\!\left[\hat{c}^{(d)}_{x,o}=c^{*(d)}_{x,o}\right].
\label{eq:r_lmh}
\end{equation}
where \(D=8\) and \(|\mathcal{O}_x|\) is the number of answer options for the current survey question. This reward directly trains the model to map option semantics into the correct region of value space, which is the transfer-critical step for unseen domains.


The \textbf{second reward} measures distributional alignment between the predicted distribution \(\hat{p}\) and the ground-truth distribution \(p^{*}\) (given by \(p_g^q\) in the constructed training episode). Following prior work, we use Jensen--Shannon divergence (JS) and define \(R_{\text{dist}} = 1 - \mathrm{JS}(\hat{p}, p^{*})\).


The \textbf{remaining two rewards} enforce schema validity. Following \citet{agarwal2025thinkinsidejsonreinforcement}, we include two binary rewards that check whether the Stage A and Stage B outputs are parseable JSON and conform to their respective schemas. These checks enforce constraints such as required keys, LMH vector length, labels restricted to \(\{\text{low},\text{medium},\text{high}\}\), inclusion of all options in the distribution, non-negativity of probabilities, and normalization within a tolerance. We denote these rewards by \(R_{\mathrm{schA}}\) and \(R_{\mathrm{schB}}\). GRPO uses the combined reward:
\begin{equation}
R=\lambda_1 R_{\mathrm{LMH}}+\lambda_2 R_{\mathrm{dist}}+\lambda_3 R_{\mathrm{schA}}+\lambda_4 R_{\mathrm{schB}}.
\label{eq:r_total}
\end{equation}
Here \(\lambda_1,\ldots,\lambda_4\) control the trade-off between semantic correctness in value space, distributional fidelity, and structural validity. Hyperparameters of GRPO (e.g., reward weights) are provided in Appendix~\ref{appendix:grpo}.

%% file: Figures/evi_da.tex
\begin{figure*}[t]    
\includegraphics[width=0.99\linewidth]{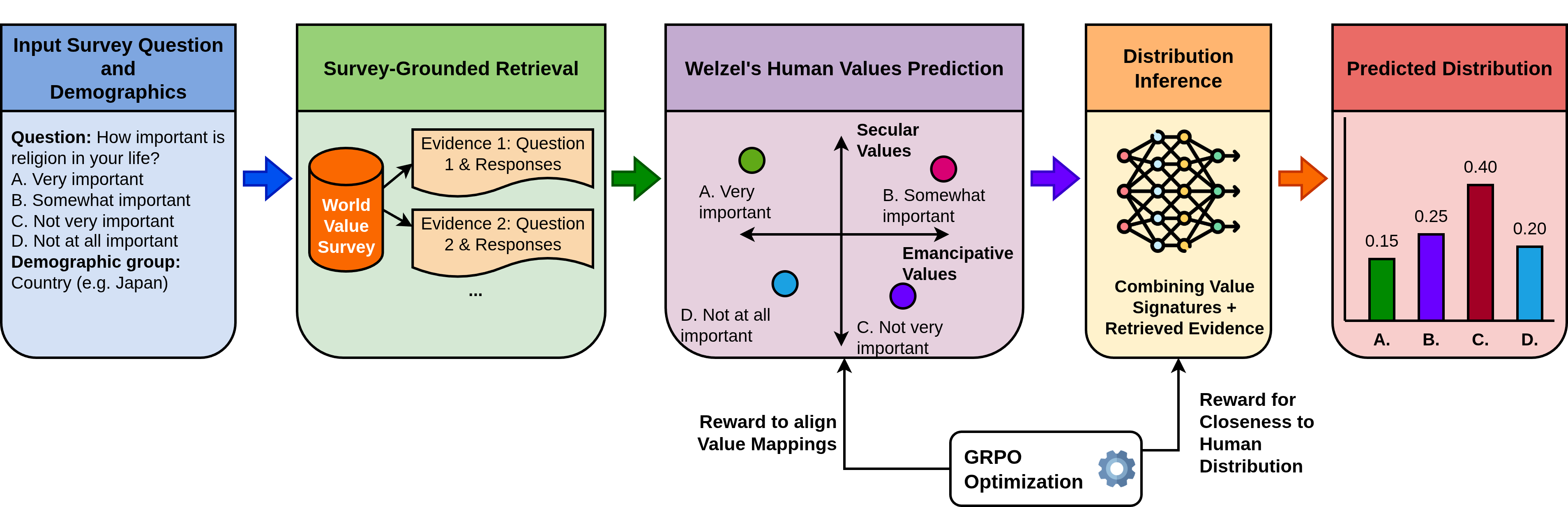}
\caption{Illustration of Evi-DA - our proposed method for distributional alignment with LLMs.}
\label{fig:evi_da}
\vspace{-1em}
\end{figure*}

%% file: latex/Experiments.tex
\section{Experiments}

\subsection{Evaluation Setup}

\paragraph{Benchmarks.} We evaluate on two survey datasets to assess both in-domain and out-of-domain performance. For in-domain evaluation, we use GlobalOpinionQA \cite{durmus2024towards} and retain only the Pew Research Center Global Attitudes Survey (GAS) portion \cite{pew_spring2024_surveydata_2025}, removing all WVS items due to duplication with our training and retrieval set. This yields 2,203 test questions. For out-of-domain evaluation, we collect survey questions released in 2025 by the Pew Research Center and YouGov to reduce potential contamination from model pretraining data. These questions cover domains not present in WVS, including \textit{Technology, Cuisine, Music, Old Age Behavior, Religious Practices,} and \textit{Gender Bias}, resulting in 200 test questions.


\paragraph{Demographic Representation.} For the evaluation of distributional alignment, demographic groups in our experiments are represented as countries, since this is the only available information in both GlobalOpinionQA and the collected out-of-domain data. 

\paragraph{Evaluation Metric.} Following previous works on distributional alignment \cite{pham-etal-2025-surveypilot, feng-etal-2024-modular}, we opt for the Jensen-Shannon Divergence score as the evaluation metric, measuring the closeness of the predicted distribution with respect to the groundtruth distribution. 



\subsection{Baselines}

To demonstrate the performance of the proposed method, we compare our framework with the following baselines:

\paragraph{Model Log-probabilities \cite{meister-etal-2025-benchmarking}.} This is the canonical distribution prediction method, in which the LLM's first token probabilities in the response are taken and assigned to each of the answer options (A, B,...).

\paragraph{Model Log-probabilities + KL \cite{cao-etal-2025-specializing}.} This baseline also uses he LLM's first token probabilities to represent distributions, then optimizes the model via a KL-divergence loss to match the human survey distribution. 

\paragraph{Opinion Sampling \cite{ferraro2024agentbasedmodellingmeetsgenerative, chuang-etal-2024-simulating}.} In opinion sampling, we follow other works on opinion synthesis and prompt the LLMs multiple times to answer survey questions. For each survey question, 10,000 responses are sampled from each LLM.

\paragraph{Verbalized Distribution \cite{meister-etal-2025-benchmarking}.} This method follows a similar approach to Evi-DA, without evidence retrieval and value mapping steps. Specificically, the LLM is instructed to provide a JSON, containing the distribution over answer options (e.g. \{A: 25\%, B: 20\%, C: 45\%, D: 10\%\}). 

\paragraph{Modular Pluralism \cite{feng-etal-2024-modular}.} Modular Pluralism is a multi-agent framework for distributional alignment, based on multi-LLM collaboration by plugging an LLM backbone to a pool of agents with profiles. For distributional alignment, each agent generates a probability distribution, and the final distribution is computed by weighted averaging each agent-provided distribution.

\paragraph{SurveyPilot \cite{pham-etal-2025-surveypilot}.} This method of distributional alignment is an agentic framework, which, instead of asking the LLM to provide the distribution like other methods, SurveyPilot collects real-world opinions from social media platforms (e.g., Reddit, X) and aggregates the comments to form the distribution. SurveyPilot is one of the leading methods for automated piloting surveys with LLM agents.

\paragraph{Evi-DA - No training.} To assess the impact of reinforcement learning, we omit GRPO training in this variant and apply Evi-DA in a zero-shot setting, where the LLM performs value mapping and distribution prediction without any additional optimization.

To evaluate both the baselines and our proposed method, we consider several open-source LLM families, including: Qwen-3 \cite{yang2025qwen3} - Qwen/Qwen3-8B, Qwen/Qwen3-14B, Qwen/Qwen3-32B; Llama-3 \cite{grattafiori2024llama3} - meta-llama/Meta-Llama-3-8B-Instruct, meta-llama/Llama-3.3-70B-Instruct; and Ministral-3-Reasoning \cite{mistralai2025mistral3release} - mistralai/Ministral-3-8B-Reasoning-2512, mistralai/Ministral-3-14B-Reasoning-2512. For each model, we use the decoding settings recommended by its authors (e.g., temperature and top-$k$). Training hyperparameters for the baselines and Evi-DA are provided in Appendix~\ref{appendix:grpo} and Appendix~\ref{appendix:baselines}.

\subsection{Main Results}

\input{Tables/tab-in-domain}

\input{Tables/tab-out-of-domain}

\input{Tables/tab-design-choice}

\input{Tables/tab-countries}

\paragraph{In-domain Evaluation.} \Cref{tab-in-domain} reports the performance of different distributional alignment methods on the in-domain test set. Even without GRPO, Evi-DA consistently outperforms prior methods. Reinforcement learning further improves performance for every LLM backbone except Ministral 14B. On average, Evi-DA achieves relative improvements over Model Log-Probabilities (44\%), Opinion Sampling (36\%), Verbalized Distribution (33\%), and SurveyPilot (30\%). In practice, a 0.1 reduction in Jensen-Shannon divergence (base-2) corresponds to a substantial redistribution of probability mass, often on the order of 10-15 percentage points across 4-5 survey options, since JSD is bounded and relates to total variation for categorical distributions \cite{corander2021jsd_tv}. Overall, these results support the use of cultural human values such as Welzel’s framework for distributional alignment of LLMs.


\paragraph{Out-of-domain Evaluation.} \Cref{tab-out-of-domain} reports results on our constructed out-of-domain benchmark. Consistent with the in-domain setting, Evi-DA outperforms other distributional alignment methods. Qwen3-32B achieves the best overall performance on this benchmark. However, for Qwen3-8B and Llama3-8B, reinforcement learning degrades performance. We attribute this to their weaker reasoning ability, which reduces accuracy in LMH signature prediction and makes it harder to use the retrieved evidence effectively when estimating the answer distribution for a new question.


\subsection{Discussions}

In this section, we explore (i) the performance of LMH signature prediction and distribution prediction across different countries and (ii) the importance of core components in Evi-DA.

\paragraph{Tasks Performance on Different Countries.} For this experiment, we analyze the performance of Evi-DA with and without reinforcement learning on 5 countries: America, Germany, France, China, and Vietnam. We only analyze the performance of Qwen3-32B for this evaluation. Table \Cref{tab-countries} shows the results of this experiment. For both of the LMH signature prediction and distribution prediction tasks, including GRPO helps to improve performance consistently across all countries. On the LMH signature prediction task, Evi-DA with reinforcement learning results in 8\% relative accuracy improvement. On the task of distribution prediction, the relative improvement is 35\%. Notably, in both tasks, the performance of Evi-DA without training on non-Western countries, such as Vietnam and China, is much lower than that of Western countries. With reinforcement learning, although there are still differences in performance across countries, it is much less biased toward America, Germany, and France. These results have justified our design choice of reward functions in GRPO for distributional alignment.

\paragraph{Importance of Core Components.} In this experiment, we measure the importance of core components in Evi-DA by removing them and evaluating the performance on the out-of-domain test set. Specifically, the following baselines are used to compare with the default settings of Evi-DA:

\begin{itemize}
    \item \textbf{Evi-DA - No training.} We remove GRPO and apply Evi-DA in a zero-shot setting.
    \item \textbf{Evi-DA - No Welzel's values.} We remove LMH signature prediction and prompt the LLM to predict the answer distribution directly using only the retrieved WVS evidence.
    \item \textbf{Evi-DA - No Evidence.} We remove evidence retrieval and prompt the LLM to predict LMH signatures for the input question without any WVS context.
\end{itemize}

\Cref{tab-design-choice} reports the results of this ablation. Removing either Welzel’s value signatures or the retrieved evidence leads to a significant drop in performance relative to the default setting. Notably, removing WVS evidence yields the largest degradation across all LLM backbones except Qwen3-14B. Overall, these results suggest that both the survey evidence and its associated Welzel value signatures are important for accurate LMH prediction and subsequent answer distribution estimation.


\paragraph{Retrieval Sensitivity Analysis.} In Appendix~\ref{appendix:exps}, we examine the sensitivity of Evi-DA to retrieval design choices by varying the number of retrieved items \(K\) and replacing BERT with stronger text encoders. We find that increasing \(K\) yields only marginal gains, so we use \(K=10\) as a good trade-off between efficiency and performance. We also find that larger retrieval encoders provide negligible improvements across all backbones on the out-of-domain test set.

%% file: Tables/tab-in-domain.tex
\begin{table*}[ht]
\centering
\resizebox{\textwidth}{!}{%
\begin{tabular}{lccccccc}
\toprule
\textbf{Method} & \textbf{Qwen3-8B} & \textbf{Qwen3-14B} & \textbf{Qwen3-32B} & \textbf{Ministral-8B} & \textbf{Ministral-14B} & \textbf{Llama3-8B} & \textbf{Average} \\
\midrule
\multicolumn{8}{c}{\cellcolor{gray!30}\texttt{Baselines}}\\
\midrule
Model Log-Probabilities  & 0.41 & 0.42 & 0.22 & 0.24 & 0.24 & 0.21 & 0.29 \\
Model Log-Probabilities + KL & 0.37 & 0.32 & 0.16 & 0.23 & 0.19 & 0.15 & 0.24 \\
Opinion Sampling         & 0.30 & 0.32 & 0.21 & 0.25 & 0.12 & 0.30 & 0.25 \\
Verbalized Distribution  & 0.26 & 0.21 & 0.20 & 0.26 & 0.31 & 0.22 & 0.24 \\
Modular Pluralism  & 0.20 & 0.21 & 0.15 & 0.18 & 0.16 & 0.19 & 0.18 \\
SurveyPilot              & 0.38 & 0.27 & 0.17 & 0.18 & 0.21 & 0.19 & 0.23 \\
\midrule
\multicolumn{8}{c}{\cellcolor{gray!30}\texttt{Proposed Method}} \\
\midrule
Evi-DA - No training & 0.21 & 0.22 & 0.14 & 0.22 & \textbf{0.11} & 0.20 & 0.18 \\
Evi-DA              & \textbf{0.19} & \textbf{0.19} & \textbf{0.11} & \textbf{0.12} & 0.16 & \textbf{0.14} & \textbf{0.16} \\
\bottomrule
\end{tabular}%
}
\caption{Results of different distributional alignment methods on the \textbf{in-domain} test set. Results are calculated on the Jensen-Shannon Divergence metric (lower is better).}
\label{tab-in-domain}
\end{table*}

%% file: Tables/tab-out-of-domain.tex
\begin{table*}[ht]
\centering
\resizebox{\textwidth}{!}{%
\begin{tabular}{lccccccc}
\toprule
\textbf{Method} & \textbf{Qwen3-8B} & \textbf{Qwen3-14B} & \textbf{Qwen3-32B} & \textbf{Ministral-8B} & \textbf{Ministral-14B} & \textbf{Llama3-8B} & \textbf{Average} \\
\midrule
\multicolumn{8}{c}{\cellcolor{gray!30}\texttt{Baselines}}\\
\midrule
Model Log-Probabilities  & 0.47 & 0.43 & 0.36 & 0.33 & 0.39 & 0.37 & 0.39 \\
Model Log-Probabilities + KL & 0.44 & 0.32 & 0.29 & 0.26 & 0.29 & \textbf{0.18} & 0.30 \\
Opinion Sampling         & 0.44 & 0.31 & 0.34 & 0.41 & 0.36 & 0.30 & 0.36 \\
Verbalized Distribution  & 0.43 & 0.39 & 0.34 & 0.29 & 0.32 & 0.38 & 0.36 \\
Modular Pluralism  & 0.32 & 0.29 & 0.22 & 0.34 & 0.34 & 0.28 & 0.29 \\
SurveyPilot              & 0.41 & 0.37 & 0.30 & 0.33 & 0.27 & 0.33 & 0.33 \\
\midrule
\multicolumn{8}{c}{\cellcolor{gray!30}\texttt{Proposed Method}} \\
\midrule
Evi-DA - No training & \textbf{0.22} & 0.30 & 0.23 & 0.20 & 0.25 & 0.23 & 0.24 \\
Evi-DA              & 0.25 & \textbf{0.27} & \textbf{0.16} & \textbf{0.17} & \textbf{0.22} & 0.28 & \textbf{0.23} \\
\bottomrule
\end{tabular}%
}
\caption{Results of different distributional alignment methods on the \textbf{out-of-domain} test set. Results are calculated on the Jensen-Shannon Divergence metric (lower is better).}
\label{tab-out-of-domain}
\end{table*}

%% file: Tables/tab-design-choice.tex
\begin{table*}[ht]
\centering
\resizebox{\textwidth}{!}{%
\begin{tabular}{lccccccc}
\toprule
\textbf{Method} & \textbf{Qwen3-8B} & \textbf{Qwen3-14B} & \textbf{Qwen3-32B} & \textbf{Ministral-8B} & \textbf{Ministral-14B} & \textbf{Llama3-8B} & \textbf{Average} \\
\midrule
Evi-DA - No training                 & \textbf{0.22} & 0.30 & 0.23 & 0.20 & 0.25 & \textbf{0.23} & 0.24 \\
Evi-DA - No Welzel's values             & 0.29 & 0.45 & 0.38 & 0.21 & 0.36 & 0.33 & 0.40 \\
Evi-DA - No Evidence & 0.48 & 0.38 & 0.46 & 0.39 & 0.55 & 0.38 & 0.44 \\
Evi-DA (Default)                     & 0.25 & \textbf{0.27} & \textbf{0.16} & \textbf{0.17} & \textbf{0.22} & 0.28 & \textbf{0.23} \\
\bottomrule
\end{tabular}%
}
\caption{Performance of Evi-DA when removing core components, measured in Jensen-Shannon Divergence (lower is better). Evaluation results calculated on the \textbf{out-of-domain} test set.}
\label{tab-design-choice}
\end{table*}

%% file: Tables/tab-countries.tex
\begin{table*}[ht]
\centering
\resizebox{\textwidth}{!}{%
\begin{tabular}{p{12cm}ccccc}
\toprule
\textbf{Method} & \textbf{America} & \textbf{Germany} & \textbf{France} & \textbf{China} & \textbf{Vietnam} \\
\midrule
\multicolumn{6}{c}{\cellcolor{gray!30}\texttt{Task: LMH Signature Prediction (Accuracy)}}\\
\midrule
Evi-DA - Without training & 0.87 & 0.89 & 0.82 & 0.78 & 0.72 \\
Evi-DA - With training    & \textbf{0.92} & \textbf{0.91} & \textbf{0.88} & \textbf{0.85} & \textbf{0.83} \\
\midrule
\multicolumn{6}{c}{\cellcolor{gray!30}\texttt{Task: Distribution Prediction (Jensen-Shannon Divergence)}}\\
\midrule
Evi-DA - Without training & 0.23 & 0.29 & 0.32 & 0.37 & 0.43 \\
Evi-DA - With training    & \textbf{0.12} & \textbf{0.18} & \textbf{0.18} & \textbf{0.22} & \textbf{0.31} \\
\bottomrule
\end{tabular}%
}
\caption{Performance of Evi-DA with and without GRPO on the tasks of LMH signature prediction and distribution prediction on different countries. Evaluation is carried out on the \textbf{out-of-domain} test set.}
\label{tab-countries}
\end{table*}

%% file: latex/RelatedWorks.tex
\section{Related Works}

\subsection{Biases in LLMs}
Recent works show LLM distributions are not culturally neutral: responses often skew toward viewpoints overrepresented in training and alignment data, frequently resembling Western value profiles \cite{naous-etal-2024-beer, Tao_2024, chiu-etal-2025-culturalbench, wang-etal-2025-proverbs}. This has been quantified by mapping model answers on survey-style instruments (e.g., WVS-like items) to real population responses across countries and demographic groups, revealing systematic gaps \cite{masoud-etal-2025-cultural}. Work such as \cite{dwivedi-etal-2025-eticor, chiu-etal-2025-culturalbench} frames cultural competence as context-sensitive judgment beyond factual knowledge, while norm-focused studies show predictable mismatches in culture- and situation-conditioned acceptability \cite{pham-etal-2025-cultureinstruct}. Most prior work evaluates individual responses or categorical correctness; we instead align population-level distributions and use a value-based intermediate representation to generalize beyond surface cues.

\subsection{Distributional and Cultural Alignment of LLMs.}

Distributional alignment complements standard alignment by requiring models to reflect the spread of opinions within a group, especially for subjective questions \cite{pluralistic, feng-etal-2024-modular}. Survey-based benchmarks such as OpinionQA \cite{opinionqa} and GlobalOpinionQA \cite{durmus2024towards} operationalize this by comparing model-elicited response distributions against real survey distributions for demographic groups and countries, finding substantial misalignment that persists even when models are prompted with group descriptors. Methodologically, existing approaches vary widely, from persona/demographic prompting \cite{feng-etal-2024-modular, pham-etal-2024-multi} to probability or sampling-based estimators and direct “verbalized distribution” outputs \cite{meister-etal-2025-benchmarking, zhang2025verbalizedsamplingmitigatemode}, with outcomes that are sensitive to elicitation format and calibration issues. Another complementary direction treats distribution estimation as an evidence-collection problem, using agentic pipelines to gather and aggregate human opinions from external sources; this improves grounding but introduces new biases and practical constraints \cite{pham-etal-2025-surveypilot}. We bridge survey-grounded targets with a transferable human-values representation and outline GRPO-style training to optimize both intermediate value prediction and final distribution fidelity.

%% file: latex/Conclusions.tex
\section{Conclusions}

In this paper, we introduced Evi-DA, an evidence-based distributional alignment method for large language models. By integrating Welzel’s values as an intermediate representation for distribution prediction and further enhancing the framework with reinforcement learning, Evi-DA improves the fidelity and robustness of population-level distribution prediction under domain and cultural shift. Through extensive evaluations, we showed that Evi-DA produces answer distributions with higher fidelity to human survey data and consistently outperforms prior distributional alignment methods, such as model logprobabilities and verbalized distributional alignment on both in-domain and out-of-domain benchmarks. Ablation studies further demonstrate that the core design choices of Evi-DA play a critical role in reducing cultural bias in distribution prediction, along with several performance analyses of alternated implementations of Evi-DA. Future work will extend Evi-DA to broader human-values frameworks and further improve generalization to unseen domains.


%% file: latex/Limitations.tex
\section*{Limitations}

While Evi-DA shows a lower Jensen-Shannon divergence than other approaches in the in-domain and out-of-domain experiments, the acceptability magnitude of the error rate from the perspective of potential users, such as social scientists, has not been addressed due to the lack of expert annotations to the tasks. Additionally, other types of human values and survey datasets outside the World Value Survey, such as surveys from PEW Research Center, are not utilized in Evi-DA training. This is due to the unavailability of clear value annotations to these data at the time of writing.

\section*{Ethical Statement}

This work studies distributional alignment - predicting how a demographic group (operationalized primarily as country) distributes responses across multiple-choice options - using publicly released survey resources (notably the World Values Survey and the Pew Research Center’s Global Attitudes Survey), plus additional publicly released survey questions collected for out-of-domain evaluation. Our experiments use group-level targets (answer distributions) and derived, de-identified value signatures computed from survey microdata; we do not run new human-subject studies, collect personal identifiers, or attempt re-identification, and we follow the respective datasets’ access conditions and citation requirements.

A central risk is that country-conditioned distributions can be misinterpreted as prescriptions (“what people should think”) or be used to reinforce stereotypes about cultures and national groups, especially when presented without uncertainty or when applied outside the survey’s population, time period, or question framing. Our intent is descriptive and research-oriented: these estimates are meant to support tasks like survey piloting, robustness testing, and population-level simulation, not to label individuals or justify consequential decisions about people. We therefore recommend that any downstream use (i) clearly communicates that outputs are probabilistic, group-level approximations, (ii) avoids high-stakes deployment (e.g., hiring, lending, political targeting), and (iii) includes domain expertise and human oversight, particularly when interpreting results about underrepresented groups or sensitive topics.

There are also methodological harms to consider. Because our evaluation and supervision are anchored to available survey ground truth, the system may overfit to how surveys operationalize questions and may compress within-country heterogeneity (e.g., regional, socioeconomic, and minority subpopulations) into a single country label, which can erase important variation. Moreover, aligning to majority response distributions can inadvertently down-weight minority viewpoints if used naively; while distributional prediction is explicitly designed to preserve disagreement relative to single-answer systems, it is still limited by what the underlying survey captures and by the demographic resolution available. We present results with these limitations in mind and frame conclusions as evidence about robustness and calibration under shift rather than definitive statements about cultures. We encourage future work on (i) stronger uncertainty quantification for distributional outputs, (ii) finer-grained demographic conditioning where ethically and legally permissible, and (iii) auditing protocols that evaluate representational harms and failure modes on sensitive domains before any real-world application.

%% file: latex/Appendix.tex
\section{Welzel's Values}
\label{appendix:welzel}

Welzel's human values framework provides a compact, interpretable way to summarize systematic cross-cultural differences in public opinion through a small number of value dimensions that recur across many domains. In this paper, we use this framework as a transferable intermediate value space that sits between question semantics and population-level response distributions: instead of asking a model to jump directly from option surface forms to a demographic-conditioned distribution, we first characterize what value profile each option expresses, and then use those value profiles to support more stable distribution inference across domains and cultures.

At the highest level, the framework is organized around two broad dimensions: \textbf{Secular Values} and \textbf{Emancipative Values}. Secular values capture the extent to which worldviews are grounded in secular-rational orientations rather than traditional authority and sacred norms, whereas emancipative values capture support for autonomy, equality, freedom of choice, and participatory voice. In our operationalization, these two higher-level dimensions are represented through \textbf{eight} sub-indexes (a fixed ordered set), where the first four correspond to the secular side and the last four correspond to the emancipative side; this yields an 8-dimensional value vector per option that can be aggregated into secular/emancipative summaries when needed.

Concretely, we use the following eight sub-indexes and their interpretations. The secular-side components are: \textbf{DEFIANCE}, capturing less deference to authority and tradition (higher indicates more defiant orientations); \textbf{DISBELIEF}, capturing lower religiosity (higher indicates more disbelief); \textbf{RELATIVISM}, capturing less moral absolutism (higher indicates more relativist moral reasoning); and \textbf{SCEPTICISM}, capturing greater skepticism toward traditional state institutions. The emancipative-side components are: \textbf{AUTONOMY}, capturing preference for independence and imagination over obedience in child-raising; \textbf{EQUALITY}, capturing support for gender equality; \textbf{CHOICE}, capturing acceptance of private-life choices (e.g., divorce, abortion, homosexuality); and \textbf{VOICE}, capturing support for free speech and people having a say. Together, these eight sub-indexes provide an interpretable coordinate system that makes intermediate commitments explicit: once an answer option is mapped into this space, it can be compared against other options (even from different domains) by the value profiles it implies, rather than by brittle lexical overlap.

For modeling, we treat each option's value representation as a vector with entries normalized to the unit interval, and we then discretize each sub-index into coarse categories to reduce noise and improve robustness in downstream reasoning. Specifically, we discretize each dimension into \textbf{low / medium / high} using fixed thresholds $\tau_1$ and $\tau_2$ (in our setup, $\tau_1 = 0.33$ and $\tau_2 = 0.67$). This discretization yields an 8-way categorical signature per option that is easy to inspect, straightforward to score automatically, and well-suited as an intermediate target for optimization and structured prediction in our pipeline.

\section{Hyperparameters}

\subsection{Evi-DA Parameters}

\subsubsection{Welzel's Values Mapping}
\label{appendix:welzel_mapping}

To perform the discretization operator in \Cref{sec:welzel}, we empirically set the values of $\tau_1$ and $\tau_2$ to 0.33 and 0.67, respectively.

\subsubsection{GRPO Parameters}
\label{appendix:grpo}

\input{Tables/tab-grpo-params}

Reward weights and core GRPO hyperparameters for Evi-DA are emperically set in \Cref{tab-grpo-param}.

\subsection{Baselines Parameters}
\label{appendix:baselines}

\paragraph{Model Log-Probabilities + KL.} We follow the default configurations from the author \cite{cao-etal-2025-specializing}, in which the learning rate for training with KL-divergence is set to 1e-4 and the batch size is set to 16. Models with this training method are optimized with AdamW.


\paragraph{SurveyPilot.} For setting up the SurveyPilot baseline, we followed the default configurations from the author \cite{pham-etal-2025-surveypilot} and set the maximum number of opinions to collect for each survey question in the benchmarks to 1,000. While more opinions can be collected, SurveyPilot takes a considerable amount of time to run, so we select this value to reduce computational cost.

\section{Additional Experiments}
\label{appendix:exps}

\subsection{Evi-DA K-Tuning}
\label{appendix:k_tuning}

In this section, we explore the sensitivity of choosing the number of retrieved survey questions $K$ when running experiments with Evi-DA on different LLM backbones. \Cref{fig:k_tuning} shows the results of this experiment. The performance of all models significantly improved when increasing $K$ from 1 to 10. With $K > 10$, the improvements are negligible and can only be observed more clearly when setting $K = 40$. Hence, to save computational resources, we set $K = 10$ in all of the main experiments.

\input{Figures/k_tuning}

\subsection{Evi-DA Retrieval Model Tuning}

We also provide experiments with different state-of-the-art embedding models other than the default configuration of Evi-DA (i.e., BERT - \texttt{google-bert/bert-base-uncased}). Results are given with different LLM backbones and different embedding models in \Cref{fig:embedding-tuning}, where we perform the experiment on the out-of-domain benchmark. Overall, there are no significant improvements to Evi-DA's performance when changing to state-of-the-art embedding models. Hence, we keep BERT as the default embedding model for retrieval to mitigate computational costs.

\input{Figures/embedding_tuning}

\section{Prompt Templates}
\label{appendix:prompts}

\input{Listings/listing_lmh_prediction}

\input{Listings/listing_distribution_prediction}

The prompt templates for LMH Signature Prediction and Distribution Prediction are provided in \Cref{listing-lmh-prediction} and \Cref{listing-distribution-prediction}, respectively.

%% file: Tables/tab-grpo-params.tex
\begin{table}[h]
    \centering
    \footnotesize
    \begin{tabular}{p{6cm}r}
        \toprule
        \textbf{Parameter} & \textbf{Value}  \\
        \midrule
        \multicolumn{2}{c}{\cellcolor{gray!30}\texttt{Reward Weights}}\\
        \midrule
        $\lambda_1$ for $R_{\text{LMH}}$ & 0.25\\
        $\lambda_2$ for $R_{\text{dist}}$ & 0.45\\
        $\lambda_3$ for $R_{\text{schemaA}}$ & 0.15\\
        $\lambda_4$ for  $R_{\text{schemaB}}$ & 0.15\\
        \midrule
        \multicolumn{2}{c}{\cellcolor{gray!30}\texttt{Core GRPO Params}}\\
        \midrule
        Group Size $G$ & 16 \\
        Clipping Param $\epsilon$ & 0.2 \\
        Epochs & 1 \\
        KL Coefficient $\beta$ & 0.04 \\
        Learning Rate & 1e-6 \\
        Batch Size & 32 \\
        \bottomrule
    \end{tabular}
    \caption{Hyperparameters of GRPO for Evi-DA.}
    \label{tab-grpo-param}
    \vspace{-1em}
\end{table}

%% file: Figures/k_tuning.tex
\begin{figure}[h]    
\includegraphics[width=0.99\columnwidth]{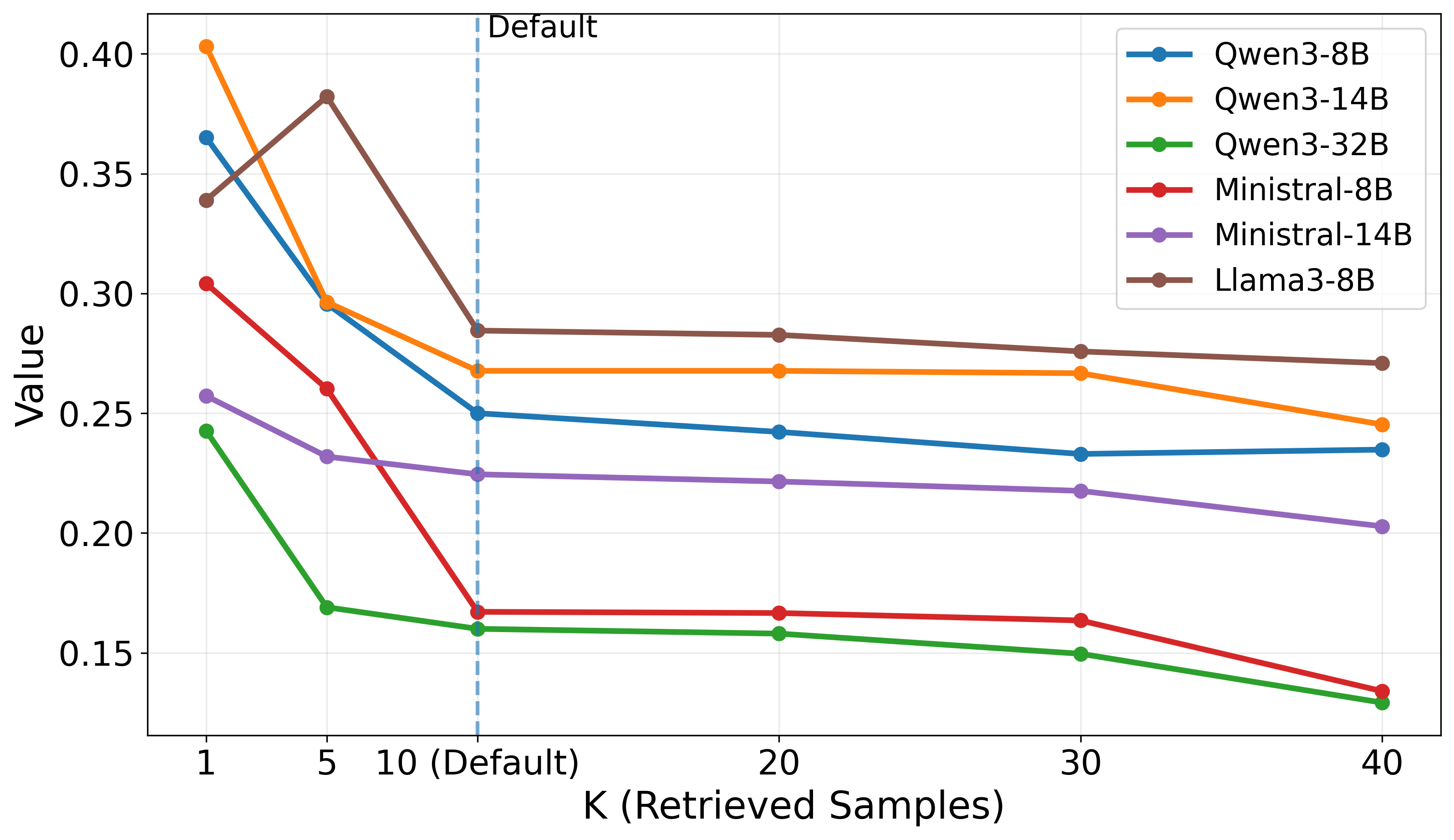}
\caption{Illustration of Evi-DA with different LLM backbones on the \textbf{out-of-domain} benchmark when changing the value $K$ for number of retrieved samples.}
\label{fig:k_tuning}
\vspace{-1em}
\end{figure}

%% file: Figures/embedding_tuning.tex
\begin{figure}[h]    
\includegraphics[width=0.99\columnwidth]{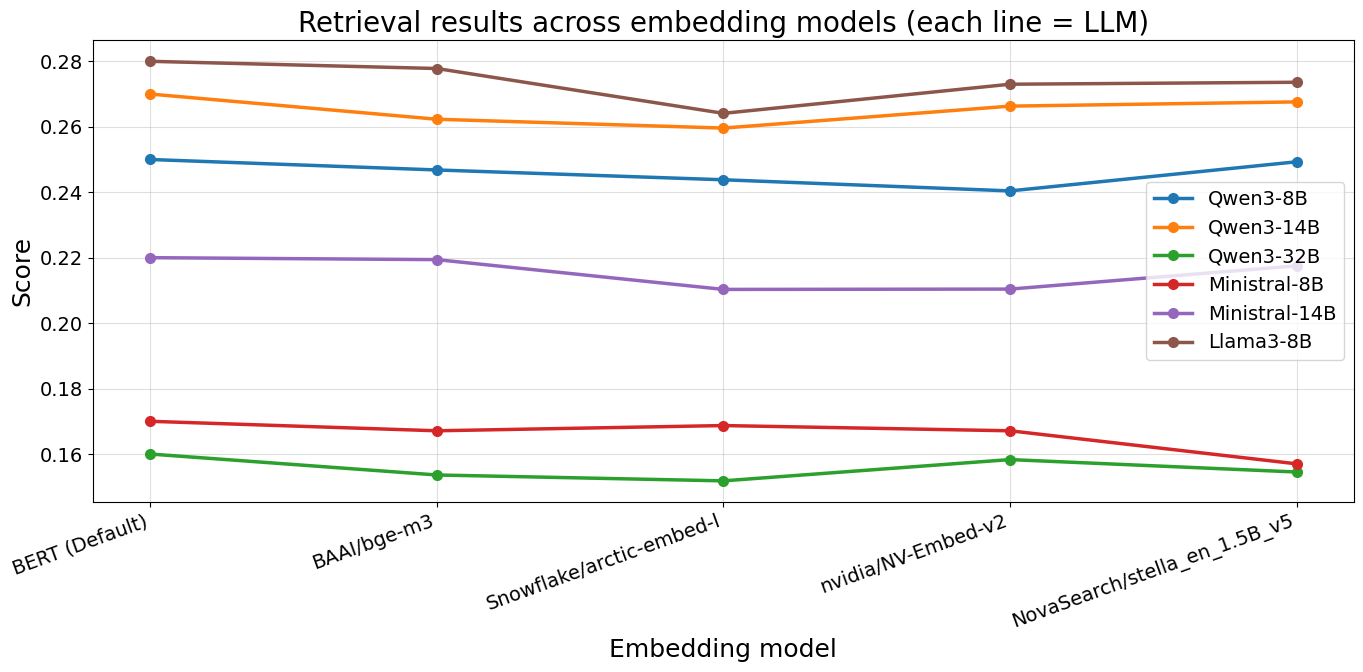}
\caption{Illustration of Evi-DA with different Embedding Models for Retrieval on the \textbf{out-of-domain} benchmark.}
\label{fig:embedding-tuning}
\vspace{-1em}
\end{figure}

%% file: Listings/listing_lmh_prediction.tex
\begin{lstlisting}[caption=Prompt template for LMH Signature Prediction, float=*, label=listing-lmh-prediction]

You are given similar WVS questions from the SAME demographic group.
Each provides:
- observed_distribution_over_labels
- label_to_subindex_LMH (typical L/M/H pattern among respondents choosing that label)

Reference questions:
{reference_questions_with_lmh}

Task:
For the input survey question, assign Welzel sub-index categories (low/medium/high) to each answer option.

Welzel sub-indexes are 8 dimensions, with discretized values including "low", "medium", and "high". Sub-index meanings:
sub_meanings = {
    "DEFIANCE": "Less deference to authority/tradition (higher = more defiant).",
    "DISBELIEF": "Lower religiosity (higher = more disbelief).",
    "RELATIVISM": "Less moral absolutism (higher = more relativist).",
    "SCEPTICISM": "More skepticism toward traditional state institutions.",
    "AUTONOMY": "Preference for independence/imagination over obedience in child-raising.",
    "EQUALITY": "Support for gender equality.",
    "CHOICE": "Acceptance of private-life choices (e.g., divorce/abortion/homosexuality).",
    "VOICE": "Support for free speech and people having a say.",
}

Sub-index order:
WELZEL_SUBINDEX_COLS = [
    "DEFIANCE", "DISBELIEF", "RELATIVISM", "SCEPTICISM",
    "AUTONOMY", "EQUALITY", "CHOICE", "VOICE",
]

Rules:
- Return ONLY JSON (no markdown).
- Use exactly the schema:
{
    "subindex_order": [
        "DEFIANCE", "DISBELIEF", "RELATIVISM", "SCEPTICISM",
        "AUTONOMY", "EQUALITY", "CHOICE", "VOICE",
    ],
    "option_profiles": [
        {"option": "<string>", "subindex_LMH": "low|medium|high"
    ],
    "notes": "<short string>",
}
- subindex_LMH entries must be one of: "low", "medium", "high"

Input survey question:
{new_question}

Answer options:
{answer_options}

Return JSON now.
\end{lstlisting}

%% file: Listings/listing_distribution_prediction.tex
\begin{lstlisting}[caption=Prompt template for Distribution Prediction, float=*, label=listing-distribution-prediction]
Context:
You are given similar WVS questions from the SAME demographic group.
Each provides:
- observed_distribution_over_labels
- label_to_subindex_LMH (typical L/M/H pattern among respondents choosing that label)

Reference questions:
{reference_questions_with_lmh}

Welzel sub-indexes are 8 dimensions, with discretized values including "low", "medium", and "high". Sub-index meanings:
sub_meanings = {
    "DEFIANCE": "Less deference to authority/tradition (higher = more defiant).",
    "DISBELIEF": "Lower religiosity (higher = more disbelief).",
    "RELATIVISM": "Less moral absolutism (higher = more relativist).",
    "SCEPTICISM": "More skepticism toward traditional state institutions.",
    "AUTONOMY": "Preference for independence/imagination over obedience in child-raising.",
    "EQUALITY": "Support for gender equality.",
    "CHOICE": "Acceptance of private-life choices (e.g., divorce/abortion/homosexuality).",
    "VOICE": "Support for free speech and people having a say.",
}

Sub-index order:
WELZEL_SUBINDEX_COLS = [
    "DEFIANCE", "DISBELIEF", "RELATIVISM", "SCEPTICISM",
    "AUTONOMY", "EQUALITY", "CHOICE", "VOICE",
]

Input survey question with Welzel's values:
{input_question_with_welzel}

Task:
For the input survey question, return a plausible probability distribution over the answer options that:
- Uses all options as keys
- Values are floats >= 0
- Sums to 1 (within rounding)
- Leans toward options whose LMH profiles better match the group target LMH profile
- Is consistent with patterns in retrieved examples (e.g., if similar LMH profiles got high probability there, mirror that)

Rules:
- Return ONLY JSON (no markdown).
- Use exactly schema: 
{
    "predicted_distribution": {"A": 0.25, "B": 0.15, "C": 0.15, "D": 0.45},
    "rationale": "<brief>",
}

Return JSON now.
\end{lstlisting}